\documentclass[11pt]{article}
\usepackage{amssymb, amsmath, amsthm}
\usepackage{graphicx,psfrag,epsf}
\usepackage{natbib}
\usepackage{multirow}
 \usepackage{subcaption}
\usepackage{graphicx}
\usepackage{algorithm}
\usepackage{algpseudocode}
\usepackage{setspace}

\def\T{{\mathrm{\scriptstyle T}}}

\newcommand{\argmin}{\operatorname*{arg \ min}}

\newcommand{\minim}{\operatorname*{minimize}}

\theoremstyle{plain}
\newtheorem{alg}{Algorithm}

\begin{document}

\def\spacingset#1{\renewcommand{\baselinestretch}%
{#1}\small\normalsize} \spacingset{1}

  \title{\bf A penalized likelihood method for\\ classification 
  with matrix-valued predictors}
  \author{Aaron J. Molstad\thanks{Corresponding author: molst029@umn.edu} \hspace{2pt} and  Adam J. Rothman\\
  School of Statistics, University of Minnesota\\}
  \date{}
  \maketitle

\bigskip
\begin{abstract}
We propose a penalized likelihood method to fit the linear 
discriminant analysis model when the predictor is matrix valued. 
We simultaneously estimate the means and the precision matrix, which we assume has a Kronecker product decomposition. 
Our penalties encourage pairs of response category mean matrices to 
have equal entries and also encourage
zeros in the precision matrix. 
To compute our estimators, we use a blockwise coordinate descent algorithm. 
To update the optimization variables corresponding to response category mean matrices, we use an alternating minimization algorithm that takes advantage of the Kronecker structure of the precision matrix. We show that our method can outperform relevant competitors in classification, even when our modeling assumptions are violated. We analyze an EEG dataset to demonstrate our method's interpretability and classification accuracy.  
\end{abstract}
\noindent%
\spacingset{1.05} 
\section{Introduction} 

We propose a method for classification when the predictor is matrix valued, e.g.
classification of hand-written letters.  
Standard vector-valued predictor classification methods, 
such as logistic regression and linear discriminant analysis, could 
be applied, but they would not take advantage of the matrix structure.

Logistic regression based methods for classification 
with a matrix-valued predictor have been proposed.  
\citet{zhou2014regularized} proposed a nuclear norm penalized likelihood estimator 
of the regression coefficient matrix 
$B_* \in \mathbb{R}^{r \times c}$ in a generalized linear model, 
where the value of the matrix predictor $x\in\mathbb{R}^{r\times c}$ enters the model through
the trace of $B_{*}^{\T} x$.
In the same setup, \citet{hung2013matrix} assumed that ${\rm vec}(B_*) = \beta_* \otimes \alpha_*$
where ${\rm vec}$ stacks the columns of its argument, 
$\alpha_* \in \mathbb{R}^{r}$, $\beta_* \in \mathbb{R}^c$, and $\otimes$ is the Kronecker product. 
This decomposition was also studied in the dimension reduction literature \citep{li2010dimension}. 

There also exist non-likelihood based methods for classification with a matrix-valued predictor.
These approaches modify Fisher's linear discriminant criterion, e.g.  
2D-LDA \citep{li20052d}, matrix discriminant analysis \citep{zhong2015matrix}, 
and penalized matrix discriminant analysis \citep{zhong2015matrix}.

We propose a penalized likelihood method for classification with a matrix-valued
predictor.  Our method estimates the parameters in the linear discriminant analysis model.
Let $x_i\in\mathbb{R}^{r\times c}$ be the measured predictor
for the $i$th subject and let $y_i\in\{1,\ldots, J\}$ be the measured categorical
response for the $i$th subject $(i=1,\ldots, n)$.  We assume that 
$(x_1, y_1), \dots, (x_n, y_n)$ are a realization of 
$n$ independent copies of $(X,Y)$ with the following distribution. 
The marginal distribution of $Y$ is defined by
$P(Y=j)=\pi_j$ $(j=1,\ldots, J)$, where the $\pi_j$'s are unknown;
and 
\begin{equation}\label{assump:Normal_Model}
 {\rm vec}(X) \mid Y=j \sim {\rm N}_{rc}\left\{ {\rm vec}\left(\mu_{*j}\right), \Sigma_*\right\}, \quad j=1,\ldots, J,
 \end{equation}
where $\mu_{*j} \in \mathbb{R}^{r \times c}$ is the unknown mean 
matrix for the $j$th response category, and $\Sigma_*$ is the unknown $rc$ by $rc$
covariance matrix.

We make the simplifying assumption that 
\begin{equation}\label{eq:Kron_decomp}
\Sigma_{*}^{-1} = \Delta_* \otimes \Phi_*,
\end{equation}
which is equivalent to $\Sigma_{*} = \Delta_{*}^{-1} \otimes \Phi_{*}^{-1}$,
where $\Phi_*$ is an unknown $r$ by $r$ precision
matrix with $\sum_{a,b} |\Phi_{*a,b}| = r$, and $\Delta_*$ is an unknown $c$ by $c$ precision matrix.
The norm condition on $\Phi_*$ is added for identifiability: see \citet{ros2016existence} for more on identifiability under \eqref{eq:Kron_decomp}.
This simplification of a covariance matrix
 makes the conditional distributions in \eqref{assump:Normal_Model} 
become matrix normal \citep{gupta1999matrix}. 
This exploits the matrix structure of the predictor
by reducing the number of parameters in the precision matrix from $O(r^2 c^2)$ to $O(r^2 + c^2)$.

Several authors have proposed and studied penalized likelihood estimators of 
$\Phi_*$ and $\Delta_*$ when $J=1$ \citep{allen2010transposable,zhang2010learning,
tsiligkaridis2012kronecker, leng2012sparse, zhou2014gemini}.

In this paper, we propose a penalized likelihood method to fit \eqref{assump:Normal_Model} 
with the assumption in \eqref{eq:Kron_decomp}.  
Our penalties encourage fitted models that can be easily interpreted by practitioners. 
We use a blockwise coordinate descent algorithm to compute our estimators.
To exploit \eqref{eq:Kron_decomp} computationally, 
we use an
alternating minimization algorithm \citep{tseng1991applications} in one of our block updates.
This algorithm scales more efficiently than other popular algorithms, which makes our method
computationally feasible for high-dimensional problems. 
We show that our algorithm has the same computational complexity order
as the unpenalized likelihood version, 
which also requires a blockwise coordinate descent algorithm \citep{dutilleul1999mle}.

\section{Penalized likelihood estimation}
\subsection{Proposed method}
Let $\mathbb{S}_{+}^m$ be the set of symmetric and positive definite $m$ by $m$ matrices.
The maximum likelihood estimators of the $\mu_{*j}$'s,  $\Phi_*$, and $\Delta_{*}$ 
minimize the function $g: (\mathbb{R}^{r \times c})^J \times \mathbb{S}_{+}^r \times \mathbb{S}_{+}^c \rightarrow \mathbb{R}$
defined by
$$
g\left(\mu, \Phi, \Delta\right) = 
\frac{1}{n}\sum_{j=1}^{J} 
\left[\sum_{i=1}^n 1(y_i = j) {\rm tr}\left\{ \Phi(x_i - \mu_j)\Delta(x_i - \mu_j)^\T\right\} \right]
- c\log {\rm det} (\Phi) - r\log {\rm det} (\Delta),
$$
where $\mu = \left(\mu_1, \dots, \mu_J\right)$. 
We propose the penalized likelihood estimators defined by
\begin{align} \label{estimator}
\left(\hat{\mu}, \hat{\Delta}, \hat{\Phi}\right) = \argmin_{(\mu, \Phi, \Delta) \in \mathcal{T}}
& \hspace{-3pt}\left\{
g(\mu, \Phi, \Delta)  + 
\lambda_1  \sum_{j < m}\| w_{j,m} \circ \left(\mu_{j} - \mu_{m}\right)\|_1 + 
\lambda_2 \|\Delta \otimes \Phi\|_1 \right\}, \\
& \hspace{50pt} \text{ subject to } \|\Phi\|_1 = r \notag
\end{align}
where $\mathcal{T} = (\mathbb{R}^{r \times c})^J \times \mathbb{S}_{+}^r \times \mathbb{S}_{+}^c$; 
$\circ$ is the Hadamard product; $\|\cdot\|_1$ is the sum of the absolute values
of the entries of its argument;   $\lambda_1$ and $\lambda_2$ are 
nonnegative tuning parameters; and
the $w_{j,m}$'s are $r$ by $c$ user-specified weight matrices. 

The first penalty in \eqref{estimator} encourages solutions for which pairs of the mean matrix estimates
have some equal entries, where this equality occurs in the same locations.  
Without the first penalty, i.e. $\lambda_1 = 0$, 
the proposed estimators of the $\mu_{*j}$'s are sample mean matrices.
If $\lambda_1 > 0$, then the proposed estimators of the $\mu_{*j}$'s are affected by 
the estimators of $\Phi_{*}$ and $\Delta_{*}$.

We recommend selecting weights similar to those prescribed by \citet{guo2010simultaneous}. We suggest using  
$$ w_{j,m}^{-1} = |\bar{x}_{j} - \bar{x}_{m}|, \quad 1 \leq j < m \leq J$$
where $\bar{x}_j = \sum_{i=1}^n 1(y_i = j) x_i$.
Alternatively, one could use weights based on $t$-test statistics 
or could use weights that incorporate prior information. 

The second penalty in \eqref{estimator} has a simple impact:
for sufficiently large values of $\lambda_2$, some of the entries in
the estimate of $\Delta_* \otimes \Phi_*$ are zero, 
which occurs if and only if either the estimate of $\Delta_*$ or the estimate of $\Phi_*$ has some zero entries. 
To encourage zeros in estimates of  $\Phi_*$ or $\Delta_*$ separately, one could 
use two separate $L_1$ penalties. Our computational algorithm can be easily adapted to accommodate this case.

The tuning parameters $\lambda_1$ and $\lambda_2$ can be chosen by minimizing the misclassification 
rate on a validation set.

\subsection{Related work}
\citet{xu2015covariance} proposed fitting the standard linear discriminant analysis model for a vector-valued predictor
by penalized likelihood.  We can express their parameter estimates in our matrix-predictor setup 
by setting the number of columns of the matrix predictor to one.  Specifically, 
with $c=1$ and $\Delta=1$, \citet{xu2015covariance} 
parameter estimates are
\begin{align} \label{eq:Guo_Xu}
\argmin_{\left(\mu, \Phi \right) \in (\mathbb{R}^{r})^J \times \mathbb{S}^{r}_+} 
\left\{ g(\mu, \Phi, 1)  
+ \lambda_1\sum_{j < m} \|w_{j,m} \circ (\mu_{j} - \mu_{m})\|_1 
 + \lambda_2 \sum_{a\neq b} |\Phi_{ab}| \right\}.
\end{align}
One could view our method as the matrix-valued predictor extension of the method of \citet{xu2015covariance}. 
\citet{guo2010simultaneous} proposed a method that solves a restricted version of \eqref{eq:Guo_Xu}, where $\Phi$ is fixed
at a diagonal matrix with pooled sample precision estimates on its diagonal.

Computationally, the algorithms proposed by \citet{xu2015covariance} and 
\citet{guo2010simultaneous} for solving \eqref{eq:Guo_Xu} suffer 
from numerical instability and do not scale efficiently for application to \eqref{estimator}. 
In our simulation studies, we compare our proposed method to several competitors, including the method of \citet{guo2010simultaneous}.  The method of \citet{xu2015covariance} is too slow computationally for the dimensions we consider, 
so we only use it in a special case when $\Sigma_*$ is known.

\section{Computation}
\subsection{Overview}
To solve \eqref{estimator}, we use a block-wise coordinate descent algorithm. 
Each block update is a convex optimization problem. In the subsequent subsections, we show that updates
for $\Phi$ and $\Delta$ can be expressed as the well-studied 
$L_1$-penalized normal likelihood precision matrix estimation problem. 
We also use an alternating minimization algorithm for the block update for $\mu.$ 
The algorithm to compute our estimator, along with a set of auxiliary functions,
 is available in the \texttt{R} package 
\texttt{MatrixLDA}, which is included in the supplemental material.

\subsection{Updates for $\Phi$ and $\Delta$}
We first derive the update for $\Phi$.  
Define ${\rm GL}(S, \tau)$ as
\begin{equation}\label{opt:graphical_lasso}
\text{GL}(S, \tau) = \argmin_{\Theta \in \mathbb{S}_+} 
\left\{ {\rm tr}(S \Theta) - \log |\Theta| + \tau \|\Theta \|_1\right\},
\end{equation}
where $S$ is some given nonnegative definite matrix and $\tau$ is a nonnegative tuning parameter. 
The optimization problem in \eqref{opt:graphical_lasso} is the $L_1$-penalized normal likelihood precision matrix estimation problem. 
Many algorithms and efficient software exist to solve \eqref{opt:graphical_lasso}: 
one good example is the graphical-lasso of \citet{friedman2008sparse}.

Let $f$ be the objective function in \eqref{estimator}. Suppose $\Delta$ and $\mu$ are fixed. The minimizer of $f$ with respect to $\Phi$ is
\begin{equation}\label{eq:Phi_udpate}
\tilde{\Phi} = \argmin_{\Phi \in \mathbb{S}^r_+} \frac{1}{n}\sum_{j=1}^{J} 
\left[\sum_{i=1}^n 1(y_i = j) {\rm tr}\left\{ \Phi(x_i - \mu_j)\Delta(x_i - \mu_j)^\T\right\} \right]
- c\log {\rm det} (\Phi) + \lambda_2 \|\Phi \otimes \Delta\|_1.
\end{equation}
Using the fact that $\| \Phi \otimes \Delta \|_1 = \|\Phi\|_1 \|\Delta\|_1$ and 
$$ \frac{1}{n}\sum_{j=1}^{J} 
 \left[\sum_{i=1}^n 1(y_i = j) {\rm tr}\left\{ \Phi(x_i - \mu_j)\Delta(x_i - \mu_j)^\T\right\} \right] = c \hspace{5pt} {\rm tr}\left\{ \Phi S_{\phi}\left(\mu, \Delta\right)\right\},$$
where 
$$ S_{\phi}(\mu, \Delta) = \frac{1}{nc} \sum_{j=1}^J
\left\{  \sum_{i=1}^n 1(y_i = j)  \left(x_i - \mu_{j}\right)\Delta
\left(x_i - \mu_{j}\right)^\T \right\},$$
we can express \eqref{eq:Phi_udpate} as
$$ \argmin_{\Phi \in \mathbb{S}^r_+}\left[ {\rm tr}\left\{ \Phi S_{\phi}(\mu, \Delta)\right\} - \log |\Phi| + \frac{\lambda_1 \| \Delta \|_1}{c} \| \Phi\|_1 \right]= \text{ GL}\left\{ S_\phi(\mu, \Delta), \frac{\lambda_1 \|\Delta\|_1}{c} \right\}.$$
After computing $\tilde{\Phi}$ with $\Delta$ fixed, we can enforce the constraint $\|\Phi\|_1 = r$ using a simple normalization: we replace $(\tilde{\Phi}, \Delta)$ with $(\bar{\Phi}, \bar{\Delta})$, where 
$$\bar{\Phi} = \frac{r}{\|\tilde{\Phi}\|_1} \tilde{\Phi}, \quad \bar{\Delta} = 
\frac{\|\tilde{\Phi}\|_1}{r} \Delta.$$
This ensures that $\|\bar{\Phi}\| = r$ without changing the objective function because $f(\mu, \Delta, \tilde{\Phi}) = f(\mu, \bar{\Delta}, \bar{\Phi})$.

Using a similar argument, the minimizer of $f$ 
with respect to $\Delta$ with $\mu$ and $\Phi$ fixed is $$ \tilde{\Delta} = {\rm GL}\left\{ S_\delta\left(\mu, \Phi\right) , \frac{\lambda_1 \|\Phi\|_1}{r}\right\},$$
where 
 $$S_{\delta}(\mu, \Phi) = \frac{1}{nr}\sum_{j=1}^J \left\{ \sum_{i=1}^n 1(y_i = j) 
\left(x_i - \mu_{j}\right)^\T \Phi \left(x_i - \mu_{j}\right)\right\}.$$

\subsection{Update for $\mu$}\label{subsec:M_update}
Let $\Delta$ and $\Phi$ be fixed.  The minimizer of $f$ with respect to $\mu$ is
\begin{equation} \argmin_{ \mu \in \mathbb{R}^{(r \times c)J} } \hspace{1pt} 
\frac{1}{n}\sum_{j=1}^{J} 
\left\{\sum_{i=1}^n \hspace{2pt} 1(y_i = j) 
{\rm tr}\left[ \Phi(x_i - \mu_j)\Delta(x_i - \mu_j)^\T \right] \right\} + \lambda_1
\sum_{j < m}\| w_{j,m} \circ \left(\mu_{j} - \mu_{m}\right)\|_1. \label{opt:AMA_original}
\end{equation}
 Special cases of \eqref{opt:AMA_original} have been solved using a majorize-minimize (MM) algorithm, where the penalty is majorized by its local-quadratic approximation at the current iterate \citep{hunter2005variable}. For example, \citet{xu2015covariance} solved \eqref{opt:AMA_original} when $c=1$ and $\Delta = 1$; and \citet{guo2010simultaneous} 
solved \eqref{opt:AMA_original} when $c=1$, $\Delta=1$, and $\Phi$ was diagonal. 
However, this MM algorithm suffers from numerical instability when iterates for $\mu_j$ and $\mu_m$ are similar from some $(j,m)$. 
Moreover, if we were to apply the MM algorithm to solve \eqref{opt:AMA_original}, 
then each iteration would 
would have worst case computational complexity $O(r^2c^2)$.

Instead of using an MM algorithm, we use an alternating minimization
algorithm \citep{tseng1991applications,chi2015splitting} to solve \eqref{opt:AMA_original}. 
Our algorithm for solving \eqref{opt:AMA_original} is more numerical stable, 
each iteration has worst case computational complexity $O(r^2c + c^2r)$ when distributed over $\max\left\{J, J(J-1)/2 \right\}$ machines, and has a quadratic rate of convergence when 
implemented with the accelerations proposed by \citet{goldstein2014fast}. Both the MM algorithm and our alternating minimization algorithm require one eigendecomposition of $\Delta$ and of $\Phi$.

Similarly to the setup of the ADMM algorithm \citep{boyd2011distributed}, we first express \eqref{opt:AMA_original} as a constrained optimization problem: 
\begin{align}\label{eq:constrained_M_update}
\minim_{(\mu, \Theta) \in \mathcal{G}}\hspace{5pt} &g(\mu, \Phi, \Delta) + \lambda_1 \sum_{j < m} \| w_{j,m} \circ \Theta_{j,m}\|\\
\text{ subject}&\text{ to } \Theta_{j,m} = \mu_j - \mu_m \quad 1 \leq j < m \leq J \notag, 
\end{align}
where $\mathcal{G} = \mathbb{R}^{(r \times c)J} \times \mathbb{R}^{(r \times c)J(J-1)/2}$ and $\Theta = \left(\Theta_{1, 2}, \dots, \Theta_{J-1, J}\right).$
The augmented Lagrangian for \eqref{eq:constrained_M_update}, using notation similar to \citet{chi2015splitting}, is
\begin{align*} 
\mathcal{F}_\rho(\mu, \Theta, \Gamma) =& g\left(\mu, \Phi, \Delta \right) +
\lambda_1 \sum_{j< m} \|w_{j,m} \circ \Theta_{j,m}\|_{1}   \\  
 & + \sum_{j<m} {\rm tr}
 \left\{\Gamma_{j,m}^\T\left(\Theta_{j,m} - \mu_{j} + \mu_{m}\right)\right\} + 
 \frac{\rho}{2}\sum_{j<m}\|\Theta_{j,m} - \mu_j + \mu_m\|_F^2, 
\end{align*}
for step size parameter $\rho > 0$ and Lagrangian variables 
$\Gamma_{j,m} \in \mathbb{R}^{r \times c}$ for $1 \leq j < m \leq J$. 
Letting the superscript $t$ denote the value of the $t$-th 
iterate of an optimization variable, the alternating minimization algorithm updates
\begin{align}
\mu^{(t+1)} & \leftarrow \argmin_{\mu \in \mathbb{R}^{(r \times c)J}} 
\mathcal{F}_0\left(\mu, \Theta^{(t)}, \Gamma^{(t)}\right), 
\label{opt: M_update} \\
\Theta^{(t+1)} & \leftarrow \argmin_{\Theta \in \mathbb{R}^{(r \times c)J(J-1)/2}} 
\mathcal{F}_\rho \left(\mu^{(t+1)}, \Theta, \Gamma^{(t)}\right), \label{opt:Theta_Update} \\
\Gamma_{j,m}^{(t+1)} & \leftarrow \Gamma_{j,m}^{(t)} + 
\rho\left(\Theta_{j,m}^{(t+1)} - \mu_{j}^{(t+1)} + 
\mu_{m}^{(t+1)}\right) \text{ for }1 \leq j < m \leq J,\notag
\end{align} 
until convergence. The ADMM algorithm modifies \eqref{opt: M_update} by using 
$\mathcal{F}_\rho$ rather than $\mathcal{F}_0$. 
The advantage of using $\mathcal{F}_0$ is that we avoid 
solving an $rc \times rc$ 
linear system of equations at complexity 
$O(r^2c^2)$ when using the Kronecker structure. Using $\mathcal{F}_0$ also allows 
the updates for $\mu_1, \dots, \mu_J$ to be computed in parallel 
with closed form solutions for each. 
Two conditions for the convergence of 
alternating minimization are that $g$ 
is strongly convex \citep{tseng1991applications}, 
which it is in our case, and that $\rho$ is sufficiently close to zero. 
We provide a computable bound on 
the step size $\rho$ to ensure convergence 
of our alternating minimization algorithm 
in the subsequent section. 

The computational advantage of alternating 
minimization over ADMM was also recognized 
by \citet{chi2015splitting} in the context 
of convex clustering. They found that the simplification 
of \eqref{opt: M_update} relative to the ADMM version yielded a substantially more efficient algorithm.

Using the first order optimality condition for \eqref{opt: M_update}, 
\begin{align}
 \mu_j^{(t+1)} &= \bar{x}_j  + \frac{1}{2\hat{\pi}_j}\Phi^{-1}\left( \sum_{ \left\{ m:  m > j \right\}} \Gamma_{j,m}^{(t)} - \sum_{\left\{ m: m < j\right\}}\Gamma_{m,j}^{(t)} \right)\Delta^{-1} \quad j=1, \dots, J,\label{eq:M_update_covs}
 \end{align} 
 where $\hat{\pi}_j = n_j/n$ for $j = 1, \dots, J$. 

The zero subgradient equation for \eqref{opt:Theta_Update} is
 \begin{equation}\label{eq:subgradient}
 \rho \Theta_{j,m}^{(t+1)} + \Gamma_{j, m}^{(t)} - \rho \left( \mu_{j}^{(t+1)} - \mu_{m}^{(t+1)} \right) + 
 \left\{ \lambda_1 w_{j,m} \circ h\left(\Theta_{j,m}^{(t+1)}\right)\right\} = 0, \quad
 \end{equation}
 where $h: \mathbb{R}^{r \times c} \to \mathbb{R}^{r \times c}$ and 
 for all $(s,t) \in \left\{ 1, \dots r\right\} \times \left\{1, \dots, c\right\}$, 
\begin{displaymath}
  \left[h(x)\right]_{s,t} = \left\{
    \begin{array}{rl}
     {\rm sign}(x_{s,t}) &: x_{s,t} \neq 0 \\
      \left[-1,1\right] &: x_{s,t} = 0
\end{array}
   \right. .
\end{displaymath} 
\citet{tibshirani1996regression}, among others, 
have shown that \eqref{eq:subgradient} can be solved using the 
soft-thresholding operator: ${\rm soft}(x, \tau) = \max(|x| - \tau, 0) {\rm sign}(x)$. 
The update for $\Theta_{j,m}$ is 
$$\Theta^{(t+1)}_{j,m} = {\rm soft}\left( \mu^{(t+1)}_{j} -
 \mu^{(t+1)}_{m} - \rho^{-1} {\Gamma^{(t)}_{j,m}}, \frac{\lambda_1}{\rho} w_{j,m}\right),$$
where ${\rm soft}$ is applied elementwise.

We use an accelerated variation of the algorithm presented in this section to solve \eqref{opt:AMA_original}. This is based on \citet{goldstein2014fast} with simple restarting rules described by \citet{o2015adaptive}. Further details about our implementation are given in the subsequent section.

\subsection{Summary}
The block-wise coordinate descent algorithm for solving \eqref{estimator} is presented in Algorithm 1. 
\begin{alg}
Given $\epsilon > 0$, $\Delta^{(0)} \in \mathbb{S}_c^+$, $\Phi^{(0)} \in \mathbb{S}_r^+$ such that $\|\Phi^{(0)}\|_1 = r$. Set $m=0:$
\begin{enumerate} 
\item[] \textit{Step 1:} Compute $\mu^{(m+1)}= \underset{\mu \in \mathbb{R}^{(r \times c)J}}{\argmin} \hspace{4pt} g\left(\mu, \Phi^{(m)}, \Delta^{(m)}\right) + \lambda_1 \sum_{j<m}\|w_{j, m} \circ \left(\mu_{j} - \mu_{m}\right)\|_1$ using the algorithm from Section \ref{subsec:M_update}. 
\item[] \textit{Step 2:} Compute $\tilde{\Delta} = {\rm GL}\left\{ S_\delta\left(\mu^{(m+1)}, \Phi^{(m)}\right) , \lambda_2\right\}$.
\item[] \textit{Step 3:} Compute $\tilde{\Phi} =  {\rm GL}\left\{ S_\phi\left(\mu^{(m+1)}, \tilde{\Delta}\right) , \frac{\lambda_2}{c}\|\tilde{\Delta}\|_1\right\}$. 
\item[] \textit{Step 4:} Compute $\Delta^{(m+1)} = \frac{\|\tilde{\Phi}\|_1}{r}\tilde{\Delta}$, $\Phi^{(m+1)} = \frac{r}{\| \tilde{\Phi}\|_1} \tilde{\Phi}$
\item[] \textit{Step 5:} If $ f\left(\mu^{(m)}, \Phi^{(m)}, \Delta^{(m)}\right) - f\left(\mu^{(m+1)}, \Phi^{(m+1)}, \Delta^{(m+1)}\right) < \epsilon |f\left(\bar{x}, \Phi^{(0)}, \Delta^{(0)}\right)|$, then stop. Otherwise, replace $m$ by $m+1$ and go to step 1. 
 \end{enumerate}
 \end{alg}

In our implementation, we set $\epsilon = 10^{-6}$. To get initial values $\Phi^{(0)}$ and $\Delta^{(0)}$, we run the maximum likelihood algorithm \citep{dutilleul1999mle} until a mild convergence tolerance is reached, and use $\Phi^{(0)} = {\rm diag}(\Phi^{\rm MLE})$ and $\Delta^{(0)} = {\rm diag}(\Delta^{\rm MLE})$ where $\left(\Phi^{\rm MLE}, \Delta^{\rm MLE}\right)$ are the final iterates. 

Let $k^{(m)}_\phi = \varphi_{\min}(\Phi^{(m)})$ and $k^{(m)}_\delta = \varphi_{\min}(\Delta^{(m)})$, where $\varphi_{\min}$ denotes the minimum eigenvalue. For the $(m+1)$th update of $\mu$, if we select the step size parameter 
\begin{equation}\label{eq:step_size_bound}
\rho^{(m+1)} \in \left(0, \left\{\min_j \left\{\hat{\pi}_j \right\} 4 k^{(m)}_\phi k^{(m)}_\delta \right\}/J \right),
\end{equation} 
then the alternating minimization algorithm converges \citep{tseng1991applications,chi2015splitting}. One can verify that \eqref{opt:AMA_original} and \eqref{eq:step_size_bound} satisfy the conditions for convergence stated in section 6.2 of the supplemental material of \citet{chi2015splitting} using an argument similar to theirs. 
 The minimum eigenvalues of $\Phi^{(m)}$ and $\Delta^{(m)}$ are positive 
as long as initializers $\Phi^{(0)}$ and $\Delta^{(0)}$ are positive definite. When $k_\delta^{(m)}$ and $k_\phi^{(m)}$ are positive, $g$ is strongly convex in $\mu$, which is required for convergence.

In practice, we find it better to use $\rho$ an order of magnitude smaller
than the upper bound in \eqref{eq:step_size_bound}, i.e., we
use $\rho^{(m+1)} =( \min_j \left\{\hat{\pi}_j \right\} 4 k^{(m)}_\phi k^{(m)}_\delta )/(10 J)$ to ensure numerical stability.
Although the step size $\rho^{(m+1)}$ 
may be small when $\Phi^{(m)}$ and $\Delta^{(m)}$ are dense, 
we find that when using accelerations and 
warm-starts, 
the small step size is not problematic. 

We use an accelerated version of the alternating minimization algorithm proposed by
\citet{goldstein2014fast}, which was also used by \citet{chi2015splitting}.
\citet{o2015adaptive} showed that acceleration restarts 
imposed after a fixed number of iterations
can decrease the number of iterations
required for convergence. 
In our implementation of the alternating minimization algorithm, 
we restart the accelerations after 200 iterations. 
We warm-start the $(m+1)$th update of $\mu$ by 
initializing the Lagrangian variables at 
their final iterates from the $m$th update. 

At convergence of the alternating minimization algorithm, zeros 
in the final iterate of ${\Theta}_{j,m}$ do not correspond to exact entrywise equality in the final iterates for $\mu_j$ and $\mu_m$. To enforce equality at the solution, we use simple thresholding. 

\subsection{Computational complexity}
Solving \eqref{estimator} with $\lambda_1 = \lambda_2 = 0$, 
i.e. maximum likelihood estimation, also requires a blockwise coordinate descent algorithm \citep{dutilleul1999mle}. 
The maximum-likelihood blockwise coordinate descent algorithm has computational complexity of 
order $O( nr^2c + nc^2r + r^3 + c^3)$. The first two terms come from computing the sample covariance matrices $S_\phi$ and $S_\delta$, and the last two terms come from inverting $S_\phi$ and $S_\delta$.

Our algorithm's computational complexity is also $O(nr^2c + nc^2r + r^3 + c^3)$.   We compute $S_\phi$ and $S_\delta$ and the graphical-lasso algorithm that we use 
is known to have worst case complexity $O(p^3)$ for a estimating a $p \times p$
precision matrix \citep{witten2011new}. 
In addition, for each $\mu$ update, we compute eigendecompositions of the iterates for $\Phi$ and $\Delta$. The alternating minimization algorithm costs $O(r^2c + c^2r)$ when implemented in parallel.

The magnitude of tuning parameters effects the computing time of our algorithm. Generally, smaller values of $\lambda_2$ take longer. 

\section{Simulation study}\label{sec:simulations}
\subsection{Models}
\begin{figure}[t]
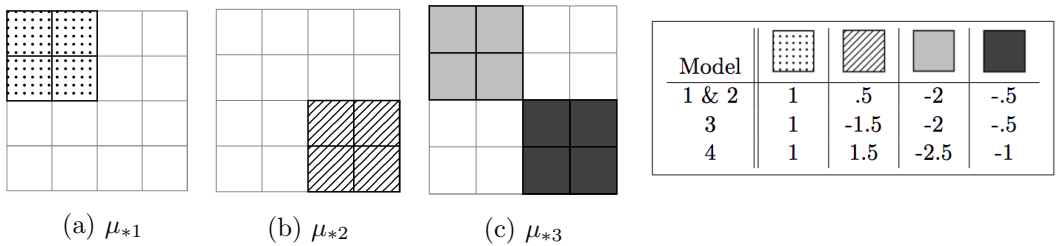

  \centering
\begin{subfigure}{.18\textwidth}
  \includegraphics[width=.95\linewidth]{Figures_MATLDA_Sims/mu1.pdf}
  \caption{$\mu_{*1}$}
  \label{fig:Table1_Fig1}
\end{subfigure}%
\begin{subfigure}{.18\textwidth}
  \centering
  \includegraphics[width=.95\linewidth]{Figures_MATLDA_Sims/mu2.pdf}
  \caption{$\mu_{*2}$}
  \label{fig:Table1_Fig2}
\end{subfigure}
\begin{subfigure}{.18\textwidth}
  \centering
  \includegraphics[width=.95\linewidth]{Figures_MATLDA_Sims/mu3.pdf}
  \caption{$\mu_{*3}$}
  \label{fig:Table1_Fig3}
\end{subfigure}
\begin{subfigure}{.40\textwidth}
  \includegraphics[width=.95\linewidth]{Figures_MATLDA_Sims/legend.pdf}
  \vspace{20pt}
\end{subfigure}
\caption{The $4 \times 4$ submatrix where $\mu_{*1}$, $\mu_{*2}$, and $\mu_{*3}$ differ. 
White corresponds to zero and the legend gives the values 
corresponding to the highlighted cells for each model.}\label{fig:Means}
\end{figure}

For 100 independent replications, we generated a realization of $n = n_{\rm train} + n_{\rm validate} + n_{\rm test}$ 
independent copies of $(X,Y)$, 
where we set $n_{\rm train} = n_{\rm validate} = 75$, and $n_{\rm test} = 1000$. 
The categorical response $Y$ has support $\left\{ 1, 2, 3\right\}$ with probabilities $\pi_{*1} = \pi_{*2} = \pi_{*3} = 1/3.$
Then $${\rm vec}\left(X\right) 
\mid Y=j  \sim {\rm N}_{rc}\left\{ {\rm vec}\left(\mu_{*j}\right), 
\Sigma_*\right\},$$ where $\mu_{*1}, \mu_{*2},$ and $\mu_{*3}$ are only different in one $4 \times 4$ submatrix, whose position is chosen randomly in each replication. 
We used multiple choices for the entries in this submatrix, which are displayed in Figure \ref{fig:Means}. 
All other mean matrix entries were set to zero. We consider four covariance models:

\begin{itemize}
\item Model 1. $\Sigma_* = \Delta_* \otimes \Phi_*$ where $\Phi_*$ has 
$(a,b)$th entry $0.7^{|a-b|}$ and $\Delta_*$ has $(c,d)$th entry $0.7 \times 1(c \neq d) + 1(c = d)$.  
\smallskip

\item Model 2. $\Sigma_* = \Delta_* \otimes \Phi_*$ where $\Phi_*$ has $(a,b)$th entry 
$0.7^{|a-b|}$ and $\Delta_*$ is block-diagonal where $\Delta_*$ can be expressed elementwise:
$$ \Delta_{c,d} = 
\left\{ 
\begin{array}{l l}
 1 & \text{  if } c=d \\
 0.7 &\text{  if } \mu_{*j,a,c} \neq \mu_{*m, a, d} \text{ for any } a \in \left\{ 1, \dots, r\right\} \text{ and } 1 \leq j < m \leq J \\
 0 & \text{  otherwise }
\end{array} 
\right. .
$$
\item Model 3. $\Sigma_*$ corresponds to the covariance model
$$
{\rm Cov}(X_{a,b}, X_{c,d} \mid Y = j)= \left\{0.5  I(b \neq d) + I(b = d)\right\} \frac{(\rho_{b}\rho_{d})^{|a-c|}}{1 - \rho_b \rho_d},
$$
where $\rho_1, \dots,\rho_c$ are $c$ 
equally spaced values between 0.5 and 0.9. The matrix $\Sigma_*$ is positive definite when $r=c$ with $c = \left\{8,16,32,64\right\}$, and when $r=32$ with $c=\left\{8, 16, 32, 64\right\}$.
\item Model 4. $\Sigma_*$ corresponds to the covariance model
$$ {\rm Cov}(X_{a,b}, X_{c,d} \mid Y=j) = 
\left\{ 
\begin{array}{l l}
 1 & \text{  if } (a,b) = (c,d) \\
 0.5 &\text{  if } \mu_{*j,a,b} \neq \mu_{*m, c, d} \text{ for any } 1 \leq j < m \leq J \\
 0 & \text{  otherwise }
\end{array} 
\right. .
$$
\end{itemize}

In Model 3, if $\rho_k = \rho$ for all $k  \in \left\{1, \dots, c\right\}$, 
then $\Sigma_*$ has the decomposition \eqref{eq:Kron_decomp} corresponding
to $\Phi_*$ with an AR(1) structure and
$\Delta_*$ with a compound symmetric structure \citep{mitchell2006likelihood}. 
However, when $\rho_k \neq \rho$, 
$\Sigma_*$ does not have decomposition \eqref{eq:Kron_decomp}:
the covariance between any two 
entries in the same row depends on the column and vice versa. 
Model 4 is the $rc-$variate 
normal model similar to the first model used 
in the simulations from \citet{xu2015covariance}. 

\subsection{Methods}
We consider the following model-based methods for fitting the linear discriminant analysis model:
\begin{itemize}
\item Bayes. The Bayes rule, i.e., $\Sigma_*$, 
$\mu_*$, and $\pi_{*j}$ known for $j = 1, \dots, J$; 
\item MN. The maximum likelihood estimator of \eqref{assump:Normal_Model} under \eqref{eq:Kron_decomp}, i.e., the matrix-normal maximum likelihood estimator; 
\item Guo. The sparse na\"ive Bayes type-estimator proposed by \citet{guo2010simultaneous} defined in Section 2.2
with tuning parameter chosen to minimize misclassification rate on the validation set; 
\item {\rm vec}-SURE. The multiclass SURE independence screening method proposed by \citet{pan2015ultrahigh}
with model sizes chosen to minimize misclassification rate on the validation set; 
\item {\rm MN}-SURE. The matrix-normal extension of the SURE 
independence screening estimator  proposed by \citet{pan2015ultrahigh} with model sizes chosen to minimize 
misclassification error on the validation set. 
 \item PMN$(\mu)$. The estimator defined by \eqref{estimator} with $\mu = \mu_*$ 
 fixed and $\lambda_2$ chosen to minimize misclassification rate on the validation set; 
 \item PMN$({\Sigma})$ / Xu($\Sigma$). The estimator defined by  
 \eqref{estimator} with $\Phi = \Phi_*$ and $\Delta = \Delta_*$ fixed 
 when $\Sigma_* = \Delta_* \otimes \Phi_*$; the estimator defined 
 by \eqref{eq:Guo_Xu} with $\hat{\Sigma} = \Sigma_*$ fixed when
  $\Sigma_* \neq \Delta_* \otimes \Phi_*$; and $\lambda_1$ chosen to minimize 
  misclassification rate on the validation set; 
  \item PMN. The estimator defined by \eqref{estimator} with tuning parameters 
  chosen by minimizing misclassification rate on the validation set.
\end{itemize}
The methods PMN$(\mu)$ and PMN$({\Sigma})$ / Xu($\Sigma$) both use some oracle information and were included to study how estimating $\mu_*$, $\Delta_*$, and $\Phi_*$ simultaneously affect classification accuracy. We refer to these method as part-oracle matrix-LDA methods. We refer to Guo and vec-SURE as vector-LDA methods; MN and MN-SURE as non-oracle matrix-LDA methods. 
MN-SURE is a matrix-normal generalization of the screening
method proposed by \citet{pan2015ultrahigh}.

Following \citet{guo2010simultaneous}, we use a validation set to select tuning parameters. 
The candidate set for tuning parameters was
 $\left\{ 2^x: x = -12, -11.5, \dots, 11.5, 12 \right\}$. 
 Candidate model sizes for vec-SURE and 
 MN-SURE were $\left\{ 0 ,1, \dots, 25\right\}$, where model size refers to the 
 number of pairwise nonzero mean differences based on thresholding.

\begin{figure}[t!]
 \centerline{\hfill\makebox[2.8in]{(a) Model 1 with $r = c$}
    \hfill\makebox[2.8in]{(b) Model 1 with $r = 32$}\hfill}
\centerline{\hfill
    \includegraphics[width=2.8in]{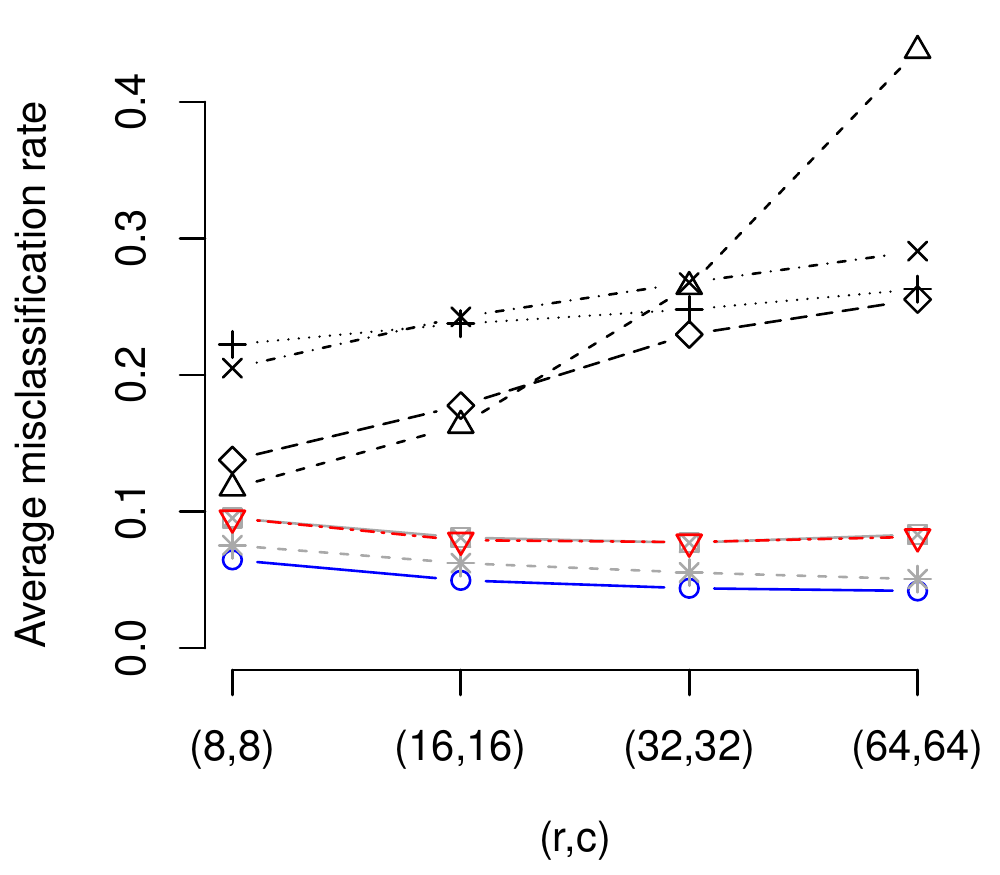}
    \hfill
   \includegraphics[width=2.8in]{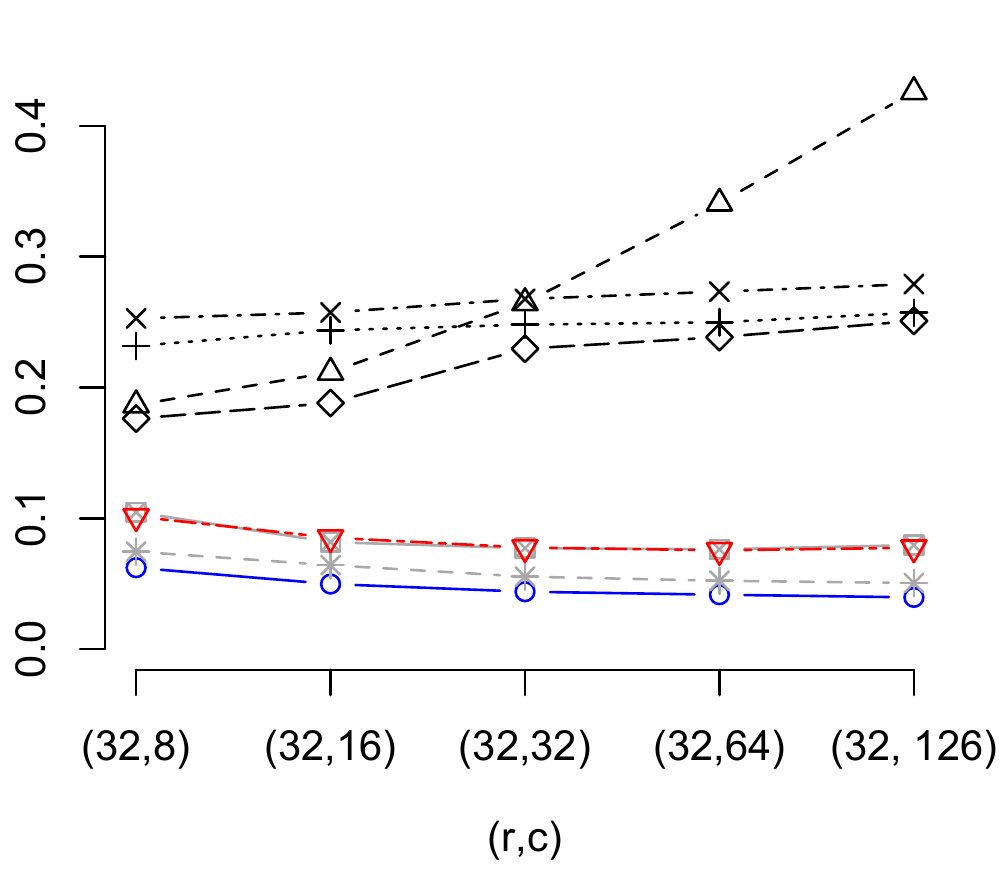}
    \hfill}  
    \centerline{\hfill\makebox[2.8in]{(c) Model 2 with $r =c $}
    \hfill\makebox[2.8in]{(d) Model 2 with $r  = 32$ }\hfill}     
\centerline{\hfill
    \includegraphics[width=2.8in]{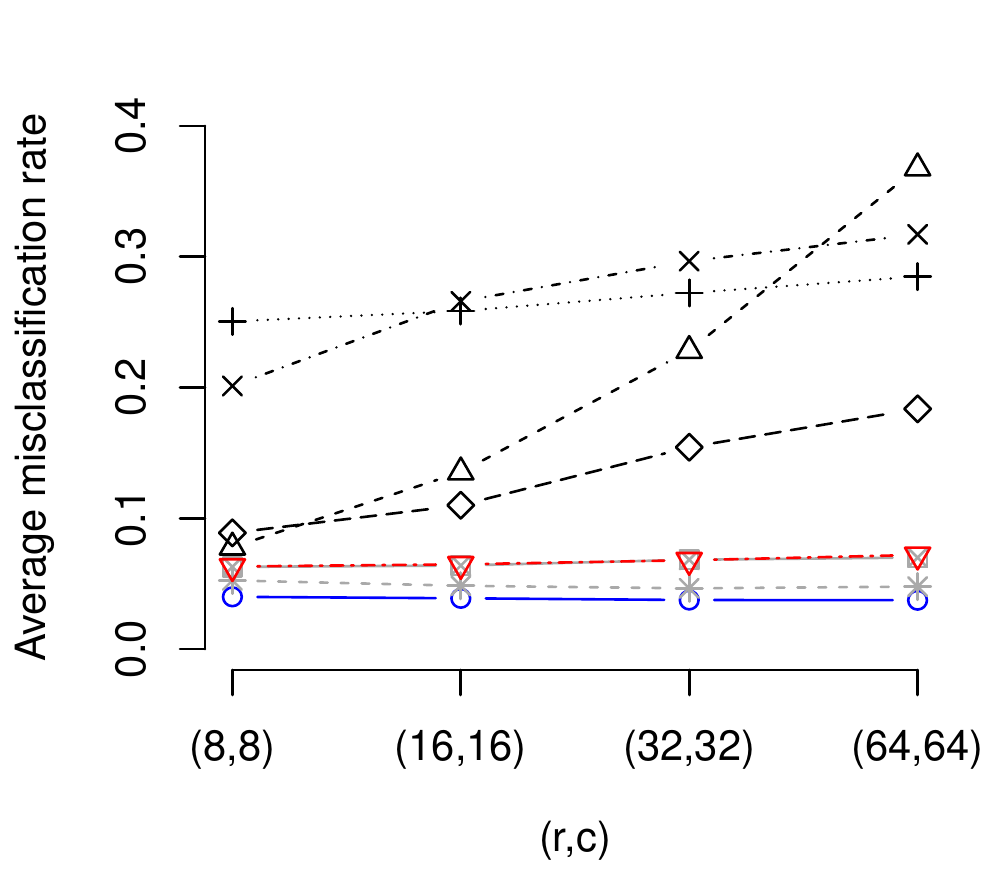}\hfill
    \hfill
    \includegraphics[width=2.8in]{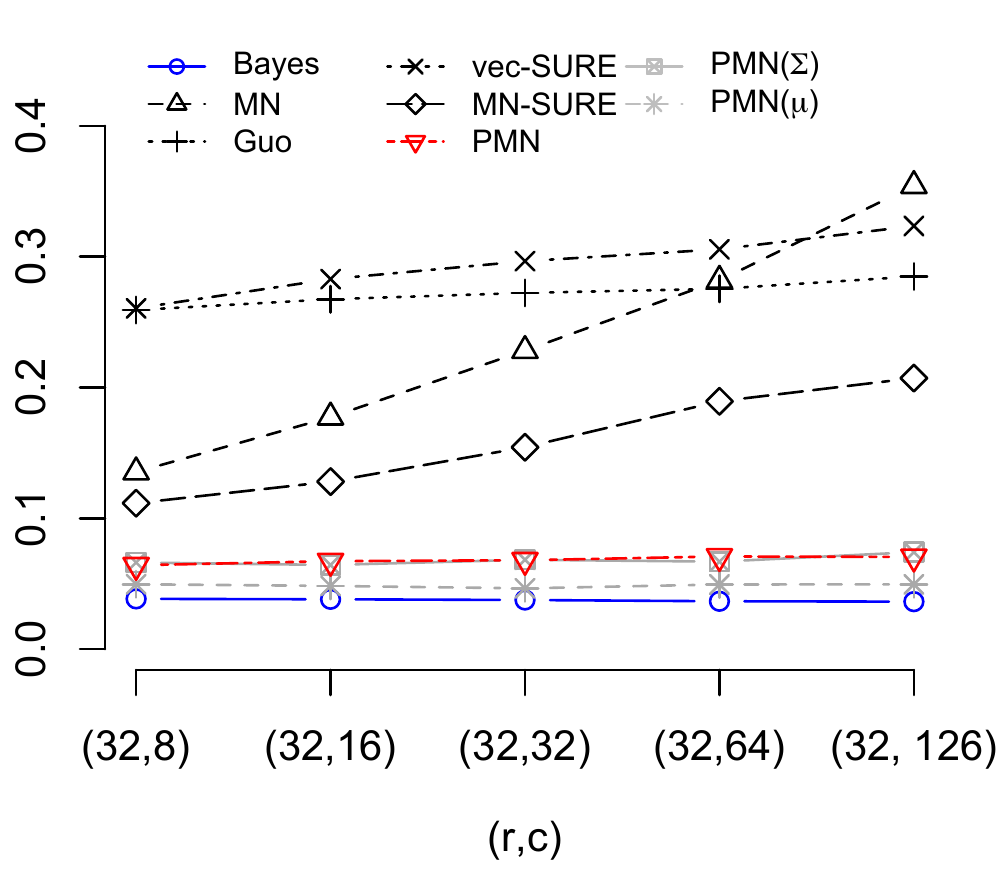}\hfill
    \hfill}       		
\caption{Misclassification rates averaged over 100 replications; (a) and (b) are for Model 1 and (c) and (d) for Model 2.}
 \label{fig:misclass_mod1_2}
\end{figure}

\begin{table}[t]
\caption{TNR/TPR percentages averaged over the 100 replications for Model 1-4. }
\label{table:TNR/TPR_mod1_2}
\scalebox{0.8}{
\fbox{
\centering
\begin{tabular}{r||cccc|cccc}
 & \multicolumn{8}{c}{Model 1 $(r,c)$}\\
Method & (8,8) & (16,16) & (32,32) & (64,64) & (32,8) & (32,16) & (32,64) & (32,126)  \\
  \hline
  Guo & 85.7/79.4 & 95.8/68.8 & 98.6/65.6 & 99.4/59.8 & 96.8/70.9 & 97.6/68.2 & 99.2/65.5 & 99.5/59.4 \\ 
  vec-SURE & 88.5/71.9 & 97.9/52.4 & 99.7/40 & 99.8/35.4 & 98.4/49.9 & 99.2/48.1 & 99.8/38.9 & 99.9/35.2 \\ 
  MN-SURE & 35.2/90.9 & 80.2/66.9 & 97.3/46.5 & 99.2/37.9 & 87.1/64.8 & 90/61.5 & 98.4/43.6 & 99.4/38.9 \\ 
  PMN($\Sigma$) & 85.9/88.2 & 94/84.1 & 98.2/78.4 & 99/81.2 & 94.1/82.6 & 96.7/83.5 & 98.7/80.6 & 99.2/74.9 \\ 
  PMN & 95.1/79.9 & 95.8/77.5 & 99/74 & 99.5/69.9 & 98.2/74.2 & 98.6/74.6 & 99.3/73.9 & 99.6/71.6 \\    
  \hline
  \hline
 &  \multicolumn{8}{c}{Model 2 $(r,c)$}\\
 & (8,8) & (16,16) & (32,32) & (64,64) & (32,8) & (32,16) & (32,64) & (32,126)  \\
\hline
 Guo  & 81.4/81.6 & 94.7/73.2 & 97.5/65.8 & 98.6/60.8 & 94.6/74.1 & 96.6/68.8 & 99/62.5 & 99/63.1 \\ 
  vec-SURE & 87.1/71.2 & 98.1/51.1 & 99.7/36.2 & 99.9/29.5 & 98.1/51.4 & 99.2/47 & 99.8/35.5 & 99.9/32.2 \\ 
  MN-SURE & 46.1/89 & 74.2/72.9 & 91.8/54 & 97.6/42.9 & 83.9/68.9 & 86.4/65.5 & 96.9/43.9 & 98.2/39.4 \\ 
  PMN($\Sigma$) & 90.9/87.5 & 94.4/86.9 & 98.8/80.6 & 99.6/84.2 & 95/86.9 & 97.4/85.5 & 99.2/82.5 & 99.4/79.9 \\ 
  PMN & 96.5/79.1 & 96.9/77.5 & 99.1/73 & 99.8/70.5 & 98.7/77.6 & 99.1/74.2 & 99.5/68.9 & 99.7/70.9 \\ 
\hline
\hline
 & \multicolumn{8}{c}{Model 3 $(r,c)$}\\
 & (8,8) & (16,16) & (32,32) & (64,64) & (32,8) & (32,16) & (32,64) & (32,126)  \\
  \hline
  Guo & 86.8/84.2 & 95.5/81.2 & 97.9/75.1 & 99.4/62.6 & 95.3/80.4 & 96/79.8 & 98.7/70.2 & ---/--- \\ 
  vec-SURE & 84.8/82.9 & 97.2/65.5 & 99.4/44.6 & 99.8/31.2 & 97.8/55.5 & 98.8/52.6 & 99.7/37 & ---/--- \\  
  MN-SURE  & 38.6/94.2 & 80.6/81.5 & 96.4/53.2 & 98.8/35.5 & 85.5/72.8 & 90.8/67.4 & 97.7/43 & ---/--- \\ 
  Xu$(\Sigma)$ & 81.4/92.1 & 93.8/90 & 96.3/86.9 & 98.5/83.9 & 91.4/87.1 & 95.7/87.9 & 97.9/87.1 & ---/--- \\ 
  PMN & 86.4/96.1 & 93.8/96 & 98.3/93 & 99.3/87.2 & 93.8/93.5 & 95.7/94.1 & 98.8/90.1 & ---/--- \\ 
  \hline
  \hline
&  \multicolumn{8}{c}{Model 4 $(r,c)$}\\
 & (8,8) & (16,16) & (32,32) & (64,64) & (32,8) & (32,16) & (32,64) & (32,126)  \\
\hline
Guo & 82.2/98.5 & 93.9/98.5 & 96.9/97.1 & 98.9/96.1 & 90.4/98.2 & 95/97.9 & 97.8/93.4 & 98.9/96 \\ 
  vec-SURE & 97.7/83.1 & 99.4/79.9 & 99.8/78.4 & 99.9/70.6 & 99.2/80.1 & 99.7/79.1 & 99.9/74.6 & 99.9/74.9 \\ 
  MN-SURE & 73.1/97.1 & 89.8/96.2 & 96/91 & 98.8/82.4 & 89.9/94.8 & 95.7/91.6 & 98.4/84.5 & 99/84.4 \\ 
  Xu$(\Sigma)$ & 87.7/99.2 & 93.3/98.4 & 97.8/96.9 & 99.5/97.1 & 92/97.8 & 96.3/96.2 & 98.8/95.8 & 99.4/96.1 \\ 
  PMN & 92.6/96.6 & 95.8/97.5 & 97.3/94.9 & 99.5/92.2 & 94.3/96.4 & 97.6/95.5 & 99.2/91.5 & 99.4/93.8 \\ 
  \end{tabular}
}}
\end{table}

 \subsection{Performance measures}
 To compare classification accuracy, we record the misclassification 
 rate on the test set for each replication. We also 
 measure identification of mean differences that are zero through both 
true positive rate (TPR) and true negative rate (TNR).  
 Let $D(\mu_*) = \left[ {\rm vec}(\mu_{*1} - \mu_{*2}), \dots,\right.\\
 \left. {\rm vec}(\mu_{*(J-1)} - \mu_{*J})  \right]$, and  
 $D(\hat{\mu}) = \left[ {\rm vec}(\hat{\mu}_{1} - \hat{\mu}_{2}), 
 \dots ,{\rm vec}(\hat{\mu}_{(J-1)} - \hat{\mu}_{J})  \right]$. 
 We define TPR as 
 $$ {\rm TPR}(\hat{\mu}, \mu_*) = 
 \frac{ \# \left\{(z,w): \left[ D(\hat{\mu})\right]_{z,w} \neq 0 \cap \left[D(\mu_*)\right]_{z,w} \neq 0\right\} }
 { \# \left\{(z,w) : \left[ D(\mu_*) \right]_{z,w} \neq 0 \right\} },$$
 where $\#$ denotes cardinality. 
We similarly define TNR as
$$ {\rm TNR}(\hat{\mu}, \mu_*) = 
 \frac{ \# \left\{(z,w): \left[ D(\hat{\mu})\right]_{z,w} = 0 \cap \left[D(\mu_*)\right]_{z,w} = 0\right\} }
 { \# \left\{(z,w) : \left[ D(\mu_*) \right]_{z,w} = 0 \right\} }.$$
 TNR and TPR together address mean difference estimation which we use as a 
 measure of variable selection for comparison to the estimator of \citet{guo2010simultaneous} and \citet{pan2015ultrahigh}.

\subsection{Results}
We display average misclassification rates for Models 1 and 2 in 
Figure \ref{fig:misclass_mod1_2}. For Model 1,
the matrix-normal maximum likelihood estimator tended to outperform 
the vector-LDA methods when $r$ and $c$ were small, but its average 
classification rate got worse as 
the dimensionality increases. The estimator proposed by \citet{guo2010simultaneous} 
performs poorly when $r$ and $c$ are small, but got worse more slowly than 
the other vector and non-oracle matrix-LDA methods. The misclassification rate 
of the Bayes rules suggests that as the dimensionality increases in Model 1, 
the optimal misclassification rate can be improved. Our method PMN had 
improved classification accuracy as both $r$ and $c$ increased and performed 
similarly to PMN($\Sigma$), which uses some oracle information. 

TPR and TNR results are displayed in Table 
\ref{table:TNR/TPR_mod1_2}. For Model 1, PMN tended to have the second highest 
TNR behind vec-SURE, but tends to have higher TPR than all competing methods 
except PMN($\Sigma$), which uses some oracle information.

\begin{figure}[t]
 \centerline{\hfill\makebox[2.8in]{(a) Model 3 with $r = c$}
    \hfill\makebox[2.8in]{(b) Model 3 with $r = 32$ }}
\centerline{\hfill
    \includegraphics[width=2.8in]{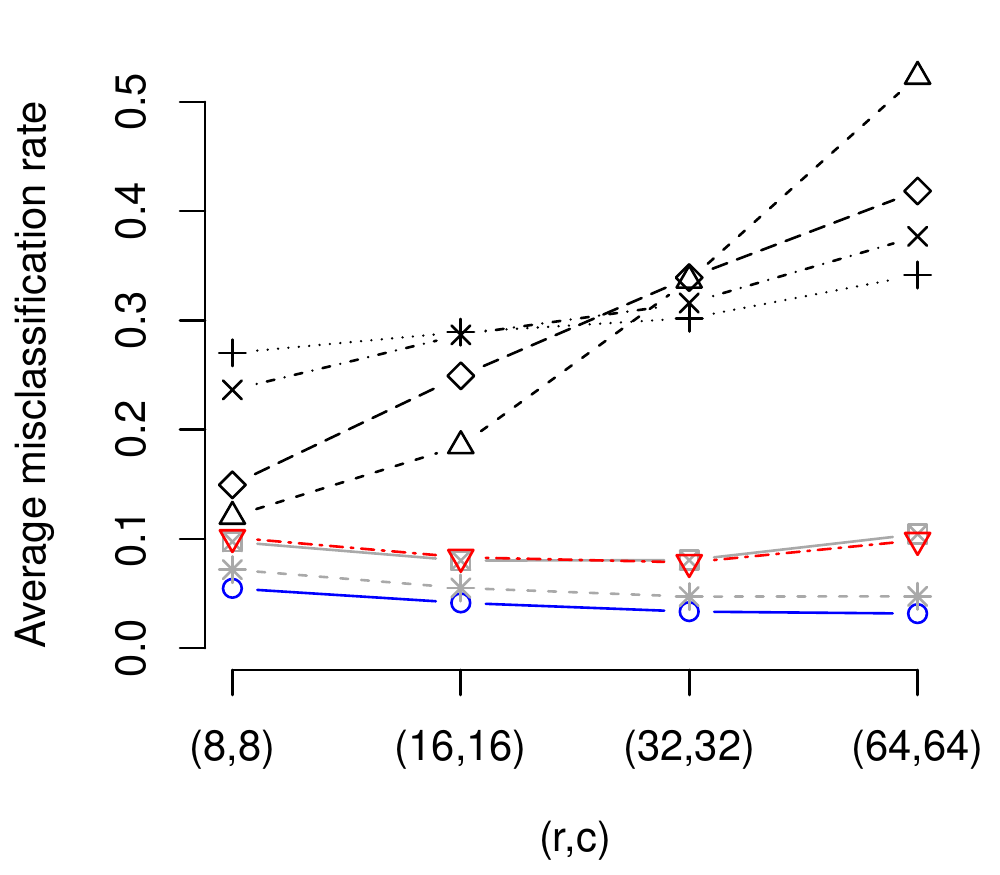}
    \hfill
   \includegraphics[width=2.8in]{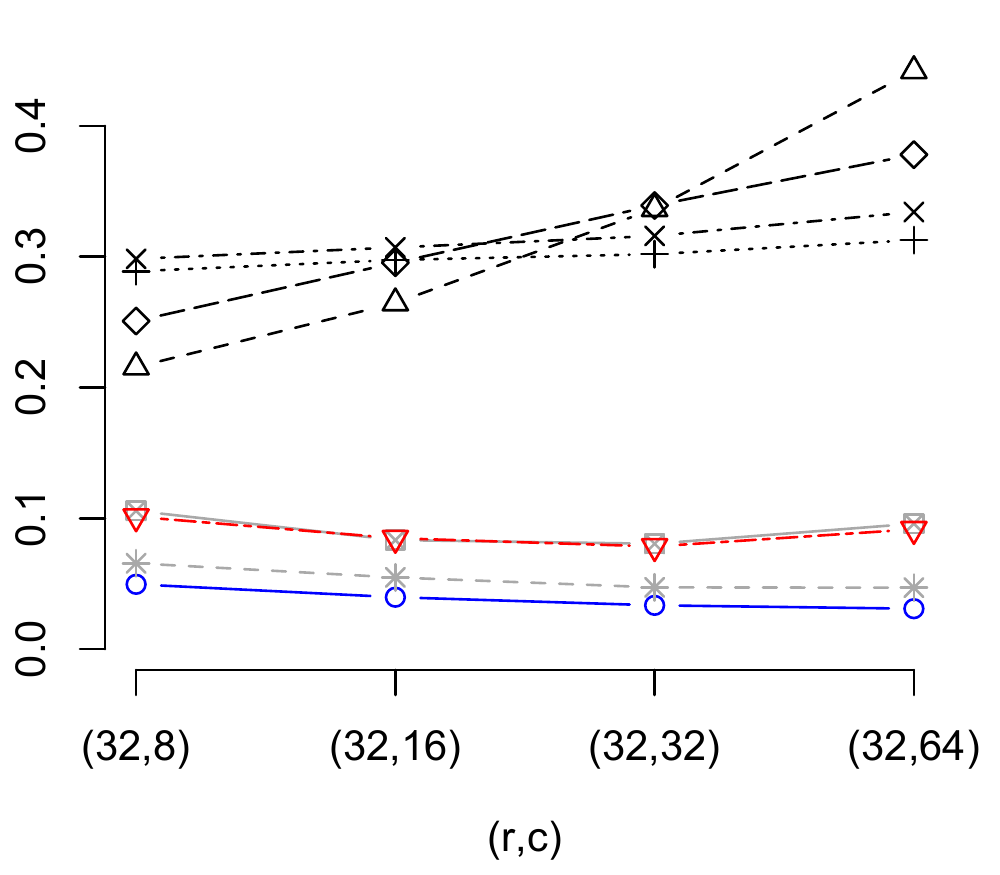}
    \hfill}  
    \centerline{\hfill\makebox[2.8in]{(c) Model 4 with $r = c $}
    \hfill\makebox[2.8in]{(d) Model 4 with $r = 32$ }\hfill}     
\centerline{\hfill
    \includegraphics[width=2.8in]{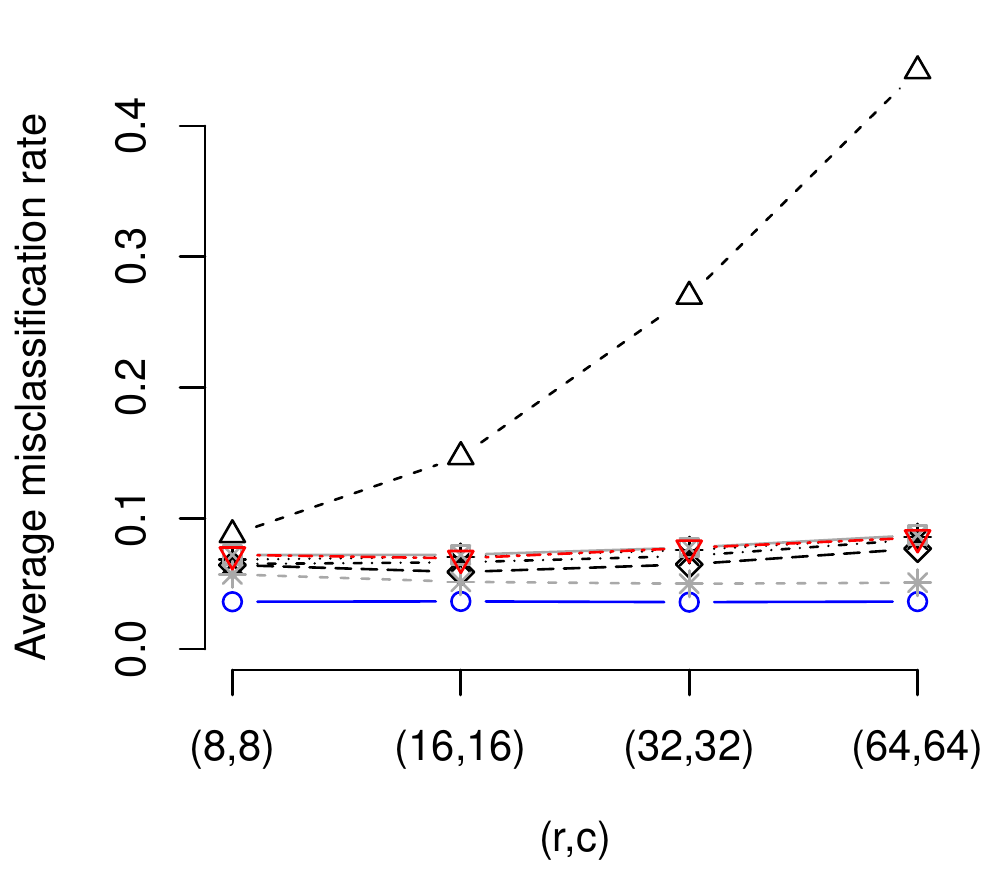}\hfill
    \hfill
    \includegraphics[width=2.8in]{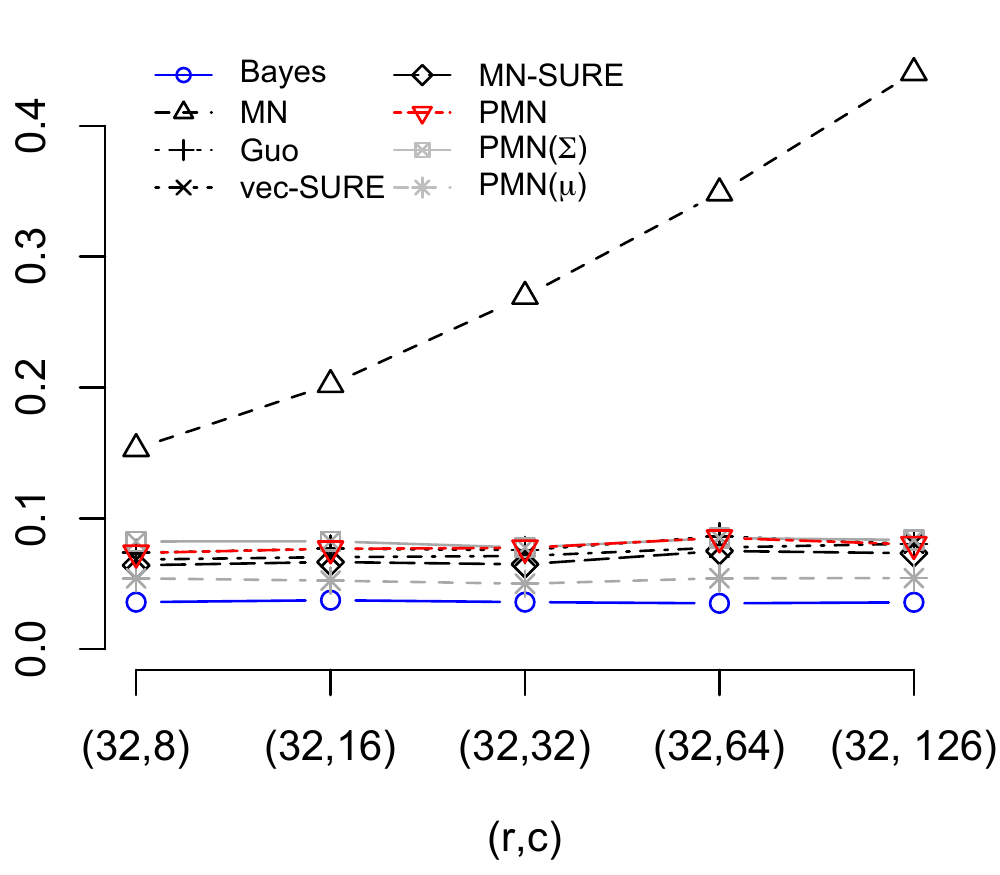}\hfill
    \hfill}       		
\caption{Misclassification rates averaged over 100 replications; (a) and (b) are for Model 3 and (c) and (d) for Model 4.}
 \label{fig:misclass_mod3_4}
\end{figure}

Results were similar for Model 2. The matrix-normal 
variation of the SURE screening estimator of \citet{pan2015ultrahigh} 
tended to perform best among the vector and non-oracle matrix-LDA methods. 
The estimator of \citet{guo2010simultaneous} got worse the
slowest amongst the vector-LDA methods. PMN performed
as well as PMN$(\Sigma)$, both of which performed more closely to PMN$(\mu$) and 
the Bayes rule than for Model 1. 

The misclassification rates for Models 3 and 4 are displayed in Figure 
\ref{fig:misclass_mod3_4}. In Model 3, 
although $\Sigma_*$ does not have the Kronecker decomposition in \eqref{eq:Kron_decomp}, 
PMN outperformed all non part-oracle 
estimators. In terms of TPR and TNR results presented in Table \ref{fig:misclass_mod3_4}, 
PMN performed similarly to 
Xu$(\Sigma)$, both of which had higher TPR than competitors and TNR similar to vec-SURE. 
This suggests that even when \eqref{eq:Kron_decomp} does not hold, 
our method can perform well in classification. 

In Model 4, PMN performed similarly to the 
vector-LDA methods. MN-SURE was the best non-oracle method, 
which suggests that \eqref{eq:Kron_decomp} may be a reasonable alternative to
na\"ive Bayes under high dimensionality. Like in Model 3, PMN performed 
similarly to Xu$(\Sigma)$ in terms of TPR and TNR.

\section{EEG data example}\label{sec:EEG_example}
We analyzed the EEG data (https://kdd.ics.uci.edu/databases/eeg/eeg.html) also studied by \citet{li2010dimension} and \citet{zhou2014regularized}. 
In the original study, 122 subjects, 77 of whom were alcoholics and 45 of whom were control, 
were exposed to stimuli while voltage was measured from $c=64$ channels on a subject's scalp at 
$r = 256$ time points. Each subject underwent 
120 trials. Each trial had one of three possible stimuli: 
single stimulus, two matched stimuli, or two unmatched stimuli.
 As in \citet{li2010dimension} and \citet{zhou2014regularized}, 
 we only analyze the single stimulus condition. 
 Because each subject underwent multiple trials under the single 
 stimulus condition, we use the within subject average over all single stimulus 
 trials as the predictor and we use whether they were alcoholic or control as the response. 

It is common to assume that \eqref{eq:Kron_decomp} holds in the 
analysis of EEG data. For example, \citet{zhou2014gemini} assumed that \eqref{eq:Kron_decomp} holds 
when analyzing a single subject from this same dataset. 
It may also be reasonable to 
assume that only a subset of channels and time point combinations are important for 
discriminating between alcoholic and control response categories. 
Thus, the primary goal of our analysis is to identify a subset of 
channels and time point combinations that help explain how
the alcoholics and controls react to the stimulus differently.

\begin{figure}[t!]
\centerline{\hfill
    \includegraphics[width=2.8in]{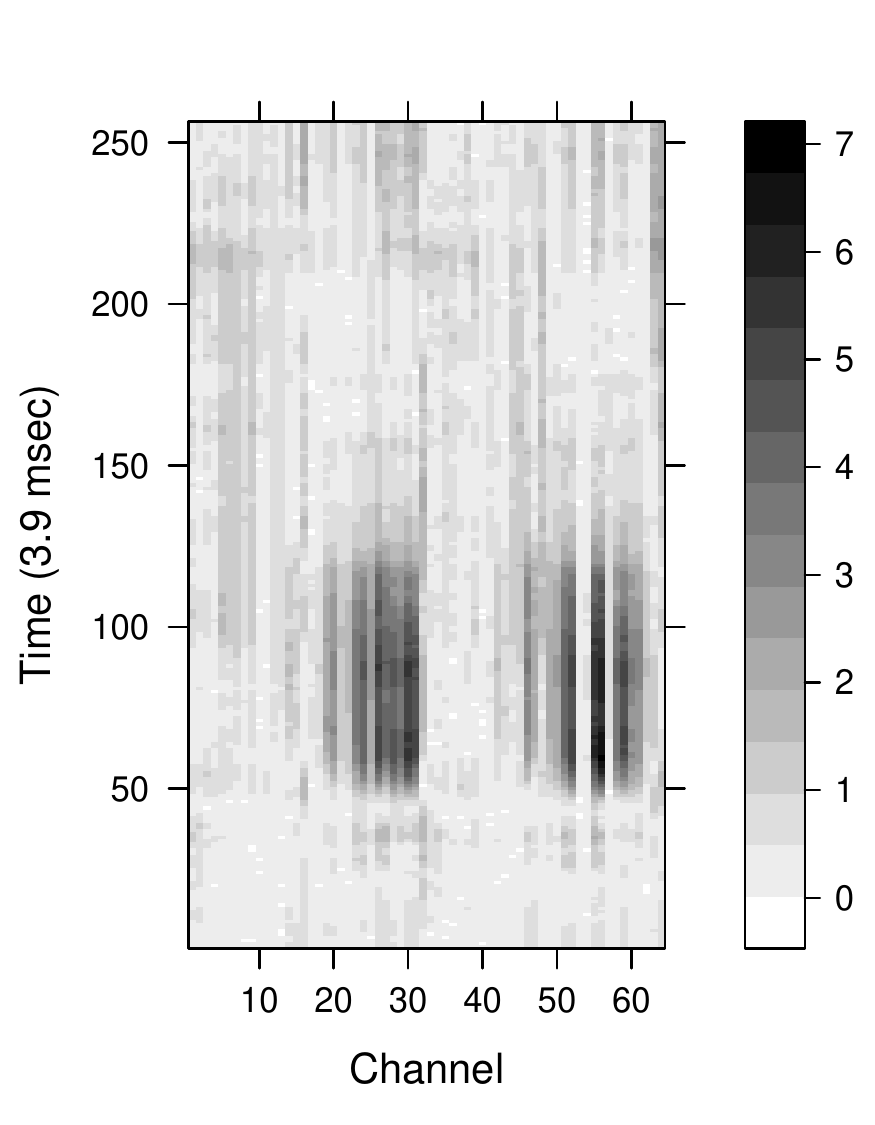}
    \hfill\includegraphics[width=2.8in]{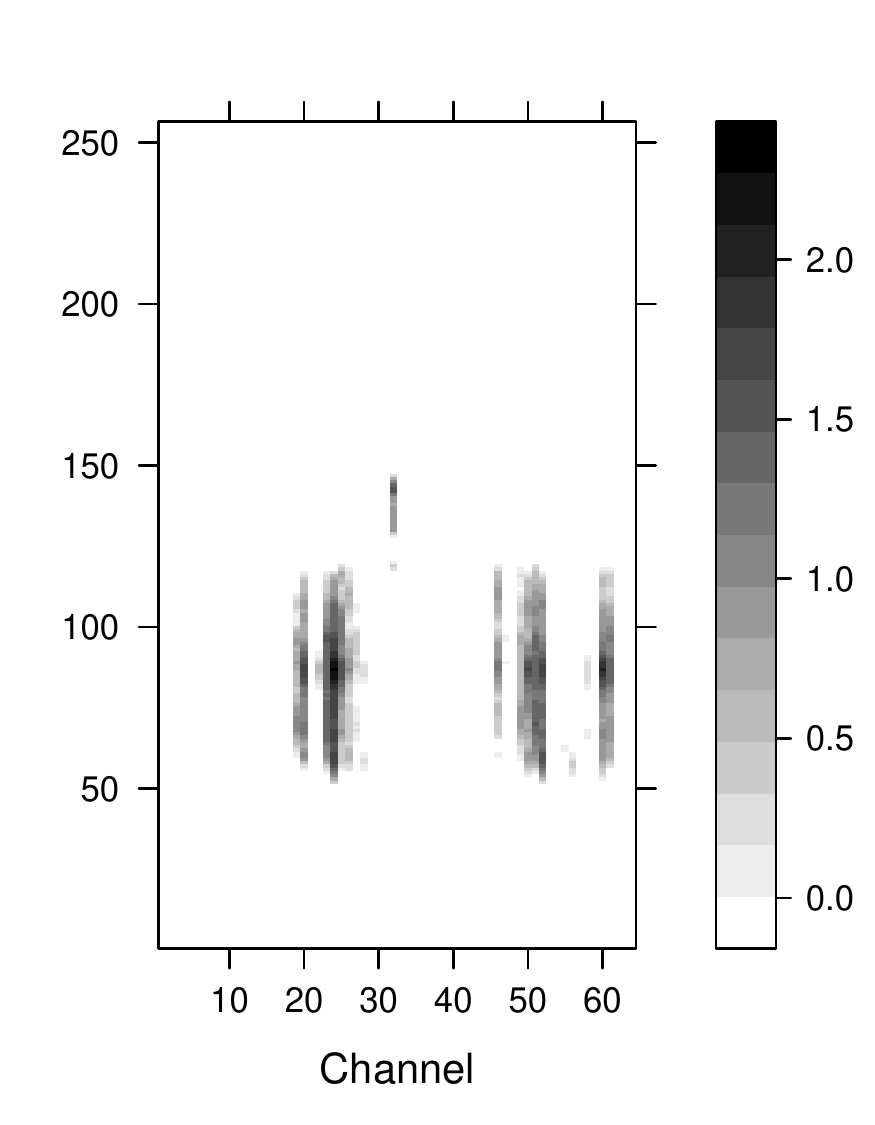}
    \hfill}  
\centerline{\hfill\makebox[2.8 in]{(a) }\hfill\makebox[2.8in]{(b)}\hfill}     		
\caption{
(a) The absolute value of the sample mean differences between 
the alcoholic and control response categories.  (b) The absolute value of the 
estimated mean differences from \eqref{estimator} based on the 
tuning parameter pair $(\lambda_1, \lambda_2) = (0.15, 5.66)$, which had leave-one-out cross-validation 
classification accuracy of 98 out of 122.}
\label{fig:mean_differences}
\end{figure}

To demonstrate our method's classification accuracy, we used the leave-one-out 
cross validation approach from \citet{li2010dimension} and \citet{zhou2014regularized}. For 
$k=1, \dots, 122$, we left out the $k$th observation and used the remaining 121 observations 
as training data. For each $k$, we selected tuning parameters for use in \eqref{estimator} by 
minimizing 5-fold cross validation misclassification error 
on the training dataset. Our method correctly classified 97 of 
122 observations. 
\citet{li2010dimension} and \citet{zhou2014regularized} reported
 correctly classifying 97 and 94 of 122, respectively. \citet{li2010dimension}
used quadratic discriminant analysis after dimension-folding of the predictors, and 
\citet{zhou2014regularized} used logistic regression 
with spectral regularization of the coefficient matrix.

To demonstrate the interpretability our fitted model, we separately fit 
\eqref{estimator} using the complete dataset. We used a tuning 
parameter pair $(\lambda_1, \lambda_2) = (0.15, 5.66)$, which had leave-one-out 
classification accuracy of 98 out of 122. The 
estimated mean difference, displayed as a heatmap in Figure 
\ref{fig:mean_differences}(b), had 15466 of 16384 entries equal to zero. 

\begin{figure}[t!]
\centerline{\hfill
    \includegraphics[width=2.6in]{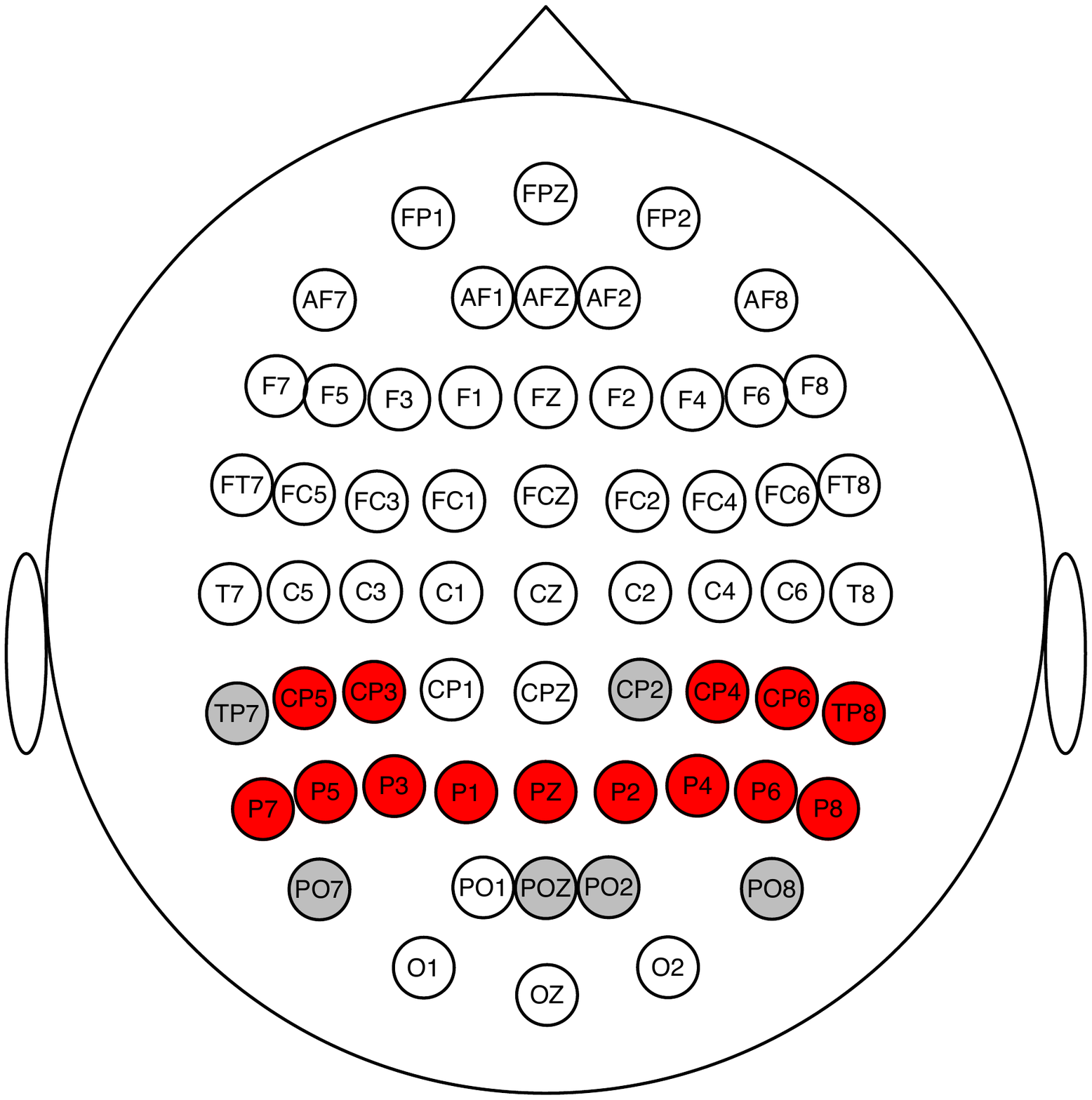}
    \hfill\includegraphics[width=2.6in]{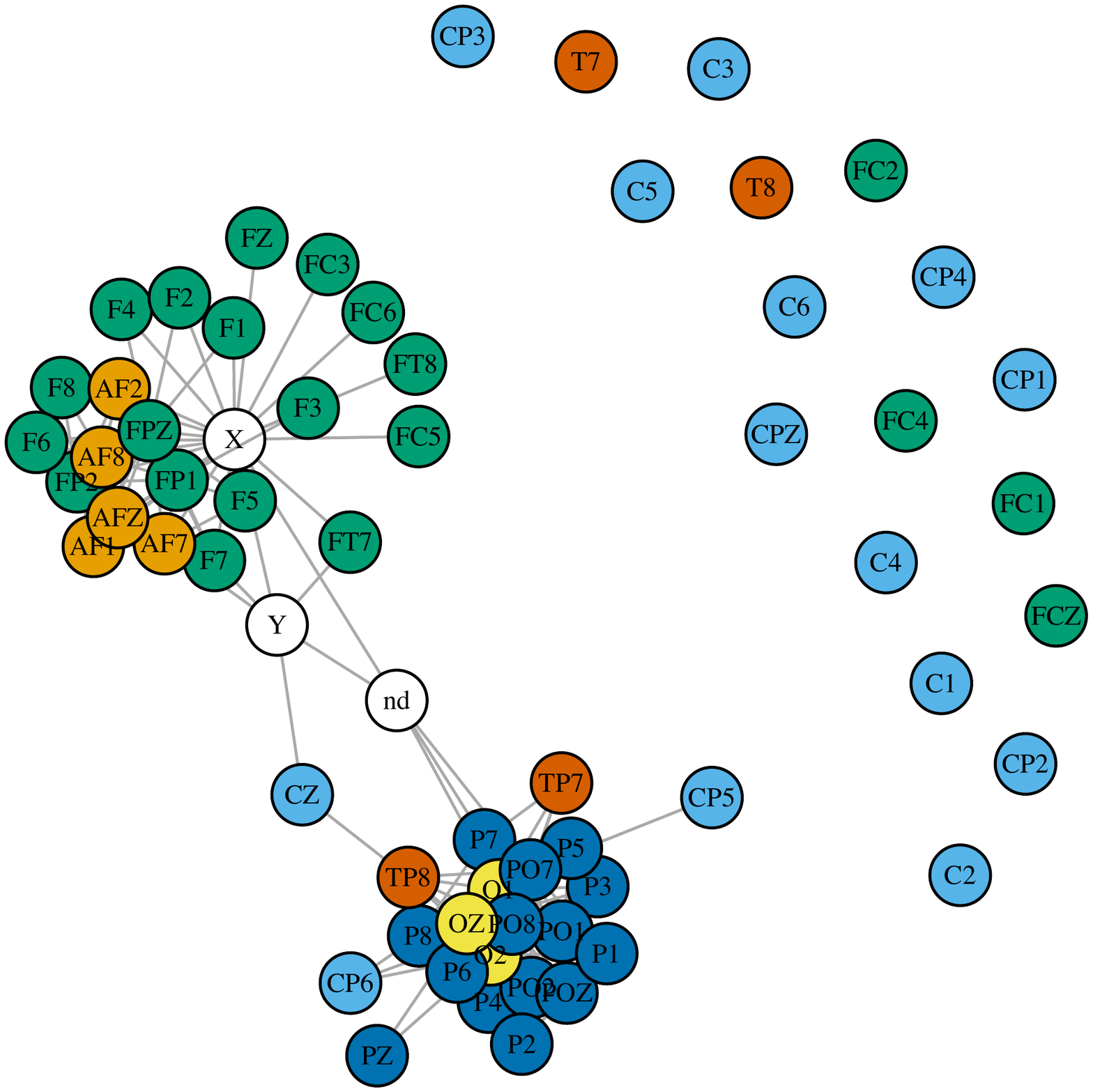}
    \hfill}  
\centerline{\hfill\makebox[2.6 in]{(a) }\hfill\makebox[2.6in]{(b)}\hfill}  
\caption{(a) An EEG cap based on the fitted model using  $(\lambda_1, \lambda_2) = (0.15, 5.66)$. 
Red channels had at 
least 20 time points estimated to have nonzero mean differences; grey channels
had less than 20 but greater than zero, whereas white channels had no nonzero mean differences.   
(b) The Gaussian precision graphical model 
associated with $\hat{\Delta}$. Colors correspond to different regions of the EEG channels; white channels are those that do not appear on the EEG cap image.}\label{fig:channels}
\end{figure}

Our fitted model
can be used to easily identify which channels and time points 
have nonzero mean differences. 
We estimated only 22 of the 64 channels to have at least one time point where 
the mean differences were nonzero, only 16 of which had at least 20 nonzero 
time points. Inspecting the estimated mean differences displayed in Figure 
\ref{fig:mean_differences}, it seems that the majority of
activity that distinguishes between the alcoholic and control subjects
takes place between the 52nd and 115th time points. 
We used the \texttt{R} package \texttt{eegkit}  \citep{helwigEEGkit}
 to display which channels had nonzero mean differences
  in Figure \ref{fig:mean_differences}a. 
Our method does not explicitly use the 
spatial structure of channels
in estimation, yet it recovered a set of 
important channels which have a natural arrangement in space.

Both $\Phi_*$ and $\Delta_*$ were estimated to be relatively sparse: 
$\hat{\Phi}$ was a diagonal matrix, while $\hat{\Delta}$ had 3676 of 
4032 off-diagonals equal to zero. 
Our estimate $\hat{\Delta}$ can be interpreted in terms of a
Gaussian precision graphical model corresponding to the conditional dependence structure of the channels. 
We display the graphical model corresponding to $\hat{\Delta}$ in Figure 
\ref{fig:channels}b. The graph structure corresponds
to the spatial arrangement of channels displayed in 
Figure \ref{fig:channels}a -- a result also
observed by \citet{zhou2014gemini}. 

\section{Extension to quadratic discriminant analysis}
Our method naturally extends to the quadratic discriminant analysis model, 
where one assumes
$$ {\rm vec}(X) \mid Y=j \sim {\rm N}_{rc}\left\{ {\rm vec}\left( \mu_{*j}\right), \Sigma_{*j}\right\} , \quad j= 1, \dots, J$$
where $\Sigma_{*j} \in \mathbb{S}_{rc}^+$ is the 
covariance matrix for the $j$th response category. 
To generalize \eqref{eq:Kron_decomp}, one can assume either
$ \text{(i) } \Sigma_{*j}^{-1} = \Delta_{*j} \otimes \Phi_{*j}$, $\text{(ii) }  
\Sigma_{*j}^{-1} = \Delta_{*j} \otimes \Phi_{*}$, or $\text{(iii) }  
\Sigma_{*j}^{-1} = \Delta_{*} \otimes \Phi_{*j}.$ 
Our algorithms can be modified to accommodate these cases. 

\section*{Acknowledgments}
This research was supported in part by the Doctoral Dissertation Fellowship from the University of Minnesota and the National Science Foundation grant DMS-1452068.

\bibliography{MatrixLDA}
\end{document}